\documentclass[twoside,11pt]{article}

\usepackage{subcaption}
\usepackage{jmlr2e}

\jmlrheading{1}{2000}{1-48}{4/00}{10/00}{Marina Meil\u{a} and Michael I. Jordan}

\ShortHeadings{On Sufficient Graphical Models}{Li and Kim}
\firstpageno{1}

\myexternaldocument{manuscript-SM}

\begin{document}
\title{On Sufficient Graphical Models}

\author{\name Bing Li \email bxl9@psu.edu \\
       \addr Department of Statistics, Pennsylvania State University\\
       326 Thomas Building, University Park, PA 16802
       \AND
       \name Kyongwon Kim \email kimk@ewha.ac.kr \\
       \addr Department of Statistics, Ewha Womans University\\
       52 Ewhayeodae-gil, Seodaemun-gu, Seoul, Republic of Korea, 03760}

\editor{}

\maketitle

\begin{abstract}
We introduce a sufficient graphical model by applying the recently developed nonlinear sufficient dimension reduction techniques to the evaluation of conditional independence. The graphical model is nonparametric in nature, as  it does not  make distributional assumptions  such as the Gaussian or copula Gaussian assumptions. However, unlike a fully nonparametric graphical model, which relies on the high-dimensional kernel to characterize conditional independence,    our graphical model is based on conditional independence given a set of sufficient predictors with a substantially reduced dimension. In this way we avoid  the curse of dimensionality that comes with a high-dimensional kernel. We develop the population-level properties,  convergence rate, and variable selection consistency of our estimate.
By simulation comparisons and an analysis of the DREAM 4 Challenge data set, we demonstrate that our method outperforms the existing methods when the Gaussian or copula Gaussian assumptions are violated, and its performance remains excellent in the high-dimensional setting.
\end{abstract}


\section{Introduction}
\label{sec;introduction}

In this paper we  propose a new nonparametric statistical graphical model, which we call the sufficient graphical model,   by {incorporating the recently developed} nonlinear sufficient dimension reduction {techniques} to the {construction of the} distribution-free graphical models.

{Let $\sten G = ( \Gamma, \ca E) $ be an undirected}  graph consist{ing} of a finite set of nodes $\Gamma=\{1, \dots, p\}$ and set of edges
$
\blue{\mathcal E \subseteq \{(i,j)\in \Gamma \times \Gamma : i \neq j \}.}
$
Since $(i,j)$ and $(j,i)$ represent the same edge in an undirected graph,  we can  assume without loss of generality that $i>j$.
A statistical  graphical model links  $\sten G$  with  a random vector $X=(X\hi1, \dots, X\hi p)$ by the conditional independence:
\begin{align}\label{eq;basicgraphical}
(i,j) \notin \mathcal E \Leftrightarrow X\hi i \indep X\hi j | X\hi{-(i,j)},
\end{align}
where
$X \hi {-(i,j)}= \{X \hi 1, \ldots, X \hi p \} \setminus \{X \hi i, X \hi j\}$, and
$A \indep B | C$ means conditional independence.  Thus, nodes $i$ and   $j$ are connected  if and only if $X \hi i$ and $X \hi j$ are  dependent given $X \hi {-(i,j)}$.
Our goal is to estimate the set $\cal E$ based on a sample $X \lo 1, \ldots, X\lo n$ of $X$.
See \citet{lauritzen1996graphical}.

One of the most popular  statistical graphical models is the Gaussian graphical model, which assumes that  $X \sim N(\mu, \Sigma)$. Under the Gaussian assumption,  conditional independence in ($\ref{eq;basicgraphical}$) is encoded in the precision matrix $\Theta = \Sigma \inv$ in the following sense
\begin{align}\label{eq;precision}
X\hi i \indep X\hi j |X\hi{-(i,j)} \Leftrightarrow \theta\lo{ij} =0,
\end{align}
where $\theta \lo {ij}$ is the $(i,j)$th entry of the precision matrix $\Theta$. By this equivalence, estimating $\ca E$ amounts  to identifying the positions of the zero entries of the precision matrix, which can be achieved by sparse estimation methods
such as the \cite{tibshirani1996regression}, \cite{fan2001variable}, and \cite{zou2006adaptive}. A variety of methods  have been developed for estimating the Gaussian graphical model, which include, for example, \cite{meinshausen2006high}, \cite{yuan2007model}, \cite{bickel2008covariance}, and \cite{peng2009partial}. See also \cite{friedman2008sparse}, \cite{guo2010pairwise}, and \cite{lam2009sparsistency}.

Since the Gaussian distribution assumption is restrictive, many recent advances  have focused on  relaxing this assumption. A main challenge in doing so is to avoid the curse of dimensionality \citep{bellman1961curse}: a straightforward nonparametric extension would resort to a high-dimensional kernel, which are known to be  ineffective.
One way to relax the Gaussian assumption without evoking a high dimensional kernel  is to use the copula Gaussian distribution, which is the approach taken by  \citet{liu2009nonparanormal}, \citet{liu2012high}, and \citet{xue2012regularized}, and is  further extended to the transelliptical model  by \citet{liu2012transelliptical}.

However,  the copula Gaussian assumption could still be restrictive: for example, if $A$ and $B$ are random variables satisfying $B=A\hi2+\epsilon$, where $A$ and $\epsilon$ are i.i.d. $N(0,1)$, then $(A,B)$ does not satisfy the copula Gaussian assumption.  To further relax the distributional assumption,  \citet{li2014additive} proposed   a new statistical relation called  {\em the additive  conditional independence}   as an alternative criterion for constructing the graphical model.  This relation has the advantage of achieving nonparametric model flexibility  without using a high-dimensional kernel, while obeying  the same set of semi-graphoid axioms that govern  the conditional independence \citep{dawid1979conditional,pearl1987logic}. See also \citet{lee2016additive} and \citet{li2017nonparametric}. Other approaches to nonparametric graphical models include \citet{fellinghauer2013stable} and \citet{voorman2013graph}.


In this paper, instead of relying on additivity  to avoid the curse of dimensionality, we apply the recently developed nonparametric sufficient dimension reduction \citep{lee2013general,li2018sufficient} to achieve this goal. The estimation proceeds in two steps: first, we use nonlinear sufficient dimension reduction to reduce  $X \hi {-(i,j)}$  to a low-dimensional random vector $U \hi {ij}$; second, we use the kernel method to construct a nonparametric graphical model based on \blue{$(X\hi i$, $X\hi j)$} and the dimension-reduced random vectors $U \hi {ij}$. The main differences between this approach and \citet{li2014additive} are, first,   we are able to retain conditional independence as the criterion for constructing the network, which is a widely accepted criterion with  a more direct interpretation, and second, we are no longer restricted by the additive structure in the graphical model. Another attractive feature of our method is  due to   the ``kernel trick'', which means its \blue{computational} complexity  depends on the sample size rather than the size of the networks.

The rest of the paper is organized as follows.   In Sections \ref{sec;basic formulation} and \ref{section:estimation-population}, we introduce the sufficient graphical model and describe its estimation method   at the population level. In Section \ref{section:estimation-sample} we lay out the detailed algorithms to implement the method. In Section \ref{section:asymptotics} we develop the asymptotic properties  such as estimation consistency, variable selection consistency, and convergence rates. In Section \ref{sec;simulation}, we conduct simulation studies to compare of our method with the existing methods. In Section \ref{sec;application}, we apply our method to the DREAM 4 Challenge gene network data set. Section \ref{sec;discussion} concludes the paper with some further discussions. Due to limited space we put all proofs and some additional results in the Supplementary Material.

\section{Sufficient graphical model}\label{sec;basic formulation}

\blue{In classical  sufficient dimension reduction, we seek the lowest dimensional subspace $\ca S$ of $\real \hi p$, such that, after projecting  $X \in \real \hi p$ on to   $\ca S$, the information about the response $Y$ is preserved; that is,  $Y \indep X | P \lo {\ca S} X$,  where $P \lo {\ca S}$ is the projection onto  $\ca S$.}  This subspace is called the central subspace, written as $\ca S \lo {Y|X}$. See, for example, \citet{li1991}, \citet{Cook1994using}, and \citet{li2018sufficient}.  \citet{li-artemiou-chiaromonte-2011} and \citet{lee2013general} extended this framework to the nonlinear setting by considering the more general problem: $Y \indep X | \ca G$, where $\ca G$ a sub-$\sigma$ field of  the $\sigma$-field generated by $X$. The class of functions in a Hilbert space that are measurable with respect to $\ca G$ is called the central class, written as $\frak S \lo {Y|X}$.  \citet{li-artemiou-chiaromonte-2011} introduced the Principal Support Vector Machine, and \citet{lee2013general} generalized the Sliced Inverse Regression \citep{li1991} and the Sliced Average Variance Estimate \citep{cook1991comment} to estimate the central class. Precursors of this theory include \citet{bach-jordan-2002}, \citet{wu2008}, and \citet{wang-yu-2008}.

To link this up with the statistical graphical model, let $(\Omega, \ca F, P)$ be a probability space, $(\Omega \lo X, \ca F \lo X)$  a Borel measurable space with $\Omega \lo X \subseteq \real \hi p$, and  $X: \Omega \to \Omega \lo X$ a random vector with distribution  $P \lo X$.
The $i$th component  of $X$ is denoted by $X \hi i$ and its range denoted by $\Omega \lo {X \hi i}$. We assume $\Omega \lo X = \Omega \lo {X \hi 1} \times \cdots \times \Omega \lo {X \hi p}$. Let $X \hi {(i,j)}=(X \hi i, X \hi j)$ and   $X \hi {-(i,j)}$ be as defined in the Introduction. Let $\sigma (X \hi {- (i,j)})$ be the $\sigma$-field generated by $X \hi {-(i,j)}$.
We assume, for each $(i,j) \in \Gamma \times \Gamma$, there is a proper sub $\sigma$-field $\ca G  \hi {-(i,j)}$ of $\sigma (X \hi {-(i,j)})$ such that
\begin{align}\label{eq:SDR assumption}
X \hi {(i,j)} \indep X \hi {-(i,j)} | \ca G \hi {-(i,j)}.
\end{align}
Without loss of generality, we assume $\ca G \hi {-(i,j)}$ is the smallest sub $\sigma$-field of $\sigma ( X \hi {-(i,j)} )$ that satisfies the above relation; that is, $\ca G \hi {-(i,j)}$ is the central $\sigma$-field for $X \hi {(i,j)}$ versus $X \hi {-(i,j)}$. \blue{There are plenty examples of joint distributions of $X$ for which the condition (\ref{eq:SDR assumption}) holds for every pair $(i,j)$: see Section \rm S10 of the Supplementary Material.}
Using the properties of conditional independence developed in \citet{dawid1979conditional} (with a detailed proof  given in  \citet{li2018sufficient}), we can show that (\ref{eq:SDR assumption}) implies the following equivalence.

\begin{theorem}\label{theorem:equivalence} If $X \hi {(i,j)} \indep X \hi {-(i,j)} | \ca G \hi {-(i,j)}$, then
\begin{align*}
X \hi i \indep X \hi j | X \hi {-(i,j)} \ \Leftrightarrow \
X \hi i \indep X \hi j | \ca G \hi {-(i,j)}.
\end{align*}
\end{theorem}

This equivalence   motivates us to use $X \hi i \indep X \hi j | \ca G \hi {-(i,j)}$ as the criterion to construct the graph $\ca G$ after performing nonlinear sufficient dimension reduction of $X \hi {(i,j)}$ versus $X \hi {-(i,j)}$ for each $(i,j) \in \Gamma \times \Gamma$, $i > j$.

\begin{definition}\label{definition:sufficient graphical model} Under condition (\ref{eq:SDR assumption}),  the graph defined by
\begin{align*}
(i,j) \notin \ca E \Leftrightarrow X \hi i \indep X \hi j | \ca G \hi {-(i,j)}
\end{align*}
is called the sufficient graphical model.
\end{definition}



\section{Estimation: population-level development}\label{section:estimation-population}

The estimation of the sufficient graphical model involves two steps: the first step is to use nonlinear sufficient dimension reduction to estimate $\ca G \hi {-(i,j)}$; the second is to construct a graph $\sten G$ based on reduced data
\begin{align*}
\{ (X \hi {(i,j)}, \ca G \hi {-(i,j)}): (i,j) \in \Gamma \times \Gamma, i > j \}.
\end{align*}
In this section we describe the two steps at the population level. To do so, we need some preliminary concepts such as the covariance operator  between two reproducing kernel Hilbert spaces,  the mean element in an reproducing kernel Hilbert spaces, the inverse of an operator,  as well as the centered reproducing kernel Hilbert spaces. These concepts are defined  in the Supplementary Material, Section \rm S1.2.  A fuller development of the related theory can be found  in \citet{li2018sufficient}. The symbols $\ran(\cdot)$ and $\overline{\ran}(\cdot)$ will be used to denote the range and the closure of the range of a linear operator.

\def\oc{\hi {\perp}}

\def\ka{\kappa}

\subsection{Step 1: Nonlinear dimension reduction}

We use the generalized sliced inverse regression \citet{lee2013general}, \citep{li2018sufficient} to perform the nonlinear dimension reduction. For each pair $(i,j) \in \Gamma \times \Gamma$, $i > j$, let $\Omega \lo {X \hi {-(i,j)}}$ be the range of $X \hi {-(i,j)}$, which is  the Cartesian product of $\Omega \lo {X \hi 1}, \ldots, \Omega \lo {X \hi p}$ with $\Omega \lo {X \hi i}$ and $\Omega \lo {X \hi j}$ removed. Let
\begin{align*}
\ka \lo X \hi {-(i,j)}: \, \Omega \lo {X \hi {-(i,j)}}  \times \Omega \lo {X \hi {-(i,j)}} \to \real
\end{align*}
be a positive semidefinite kernel.
Let $\sten H \lo X \hi {-(i,j)}$ be the centered reproducing kernel Hilbert space generated by $\ka \lo X  \hi { {-(i,j)}}$. Let $\Omega \lo {X \hi {(i,j)}}$, $\ka \lo X \hi {{(i,j)}}$, and $\sten H \lo X \hi {(i,j)}$ be the similar objects defined for $X \hi {(i,j)}$.

\begin{assumption}\label{assumption:finite expectation of kernel}
\begin{align*}
E[ \ka \lo X \hi {-(i,j)} ( X \hi {-(i,j)}, X \hi {-(i,j)} ) ]< \infty, \quad E[ \ka \lo X \hi {(i,j)} ( X \hi {(i,j)}, X \hi {(i,j)} )] < \infty.
\end{align*}
\end{assumption}
\blue{This is a very mild assumption that is satisfied by most kernels. }
Under this assumption, the following covariance operators are well defined:
\begin{align*}
\Sigma \lo {X \hi {-(i,j)} X \hi {(i,j)}}:  \sten H \lo X \hi {(i,j)} \to \sten H \lo X \hi {-(i,j)}, \quad
\Sigma  \lo {X \hi {-(i,j)} X \hi {-(i,j)}}:  \sten H \lo X \hi {-(i,j)} \to \sten H \lo X \hi {-(i,j)}.
\end{align*}
For the formal definition of the covariance operator, see S1.2. Next, we introduce the regression operator from $\sten H \lo X \hi {(i,j)}$ to $\sten H \lo X \hi {-(i,j)}$. For this purpose we need to make the following assumption.
\begin{assumption}\label{assumption:ran in ran}
$\ran ( \Sigma \lo {X \hi {-(i,j)} X \hi {(i,j)} } ) \subseteq \ran (  \Sigma \lo {X \hi {-(i,j)} X \hi {-(i,j)} } )$.
\end{assumption}
As argued in \citet{li2018sufficient}, this   assumption can be interpreted as a type of collective  smoothness  in the relation between $X \hi {(i,j)}$ and $X \hi {-(i,j)}$: intuitively, it requires the operator $\Sigma  \lo {X \hi {-(i,j)} X \hi {(i,j)}}$ sends all the  input functions to the low-frequency domain of the operator $\Sigma  \lo {X \hi {-(i,j)} X \hi {-(i,j)}}$. Under \blue{Assumption \ref{assumption:ran in ran}}, the linear operator
\begin{align*}
R \lo {X \hi {-(i,j)} X \hi {(i,j)}}=\Sigma \lo {X \hi {-(i,j)} X \hi {-(i,j)} } \inv\Sigma \lo {X \hi {-(i,j)} X \hi {(i,j)} }
\end{align*}
is defined, and we call it the regression operator from $\sten H \lo X \hi {(i,j)}$ to $\sten H \lo X \hi {-(i,j)}$. The meaning of the inverse
$\Sigma \lo {X \hi {-(i,j)} X \hi {-(i,j)} } \inv$ is defined in Section \rm S1.2 in the Supplementary Material.
The regression operator in this form was formally defined in \citet{lee2016}, but earlier forms existed in \citet{fukumizu2004dimensionality}; see also \citet{li2018linear}.
\begin{assumption} $R \lo {X \hi {-(i,j)} X \hi {(i,j)}}$ is a finite-rank operator, with rank $d \lo {ij}$.
\end{assumption}
\blue{Intuitively, this assumption  means that $R \lo {X \hi {-(i,j)}X \hi {(i,j)}}$ filters out the high frequency functions of $X \hi {(i,j)}$, so that, for any $f \in \sten H \hi {\, (i,j)}$, $R \lo {X \hi {-(i,j)}X \hi {(i,j)}} f$ is relatively smooth. It will be violated, for example,  if  one can  find an $f \in \sten H \hi {\, (i,j)}$ that makes $R \lo {X \hi {-(i,j)}X \hi {(i,j)}} f$ arbitrarily choppy.}
The regression operator plays a crucial role  in nonlinear sufficient dimension reduction.  Let  $L \lo 2 ( P \lo {X \hi {-(i,j)}} )$  be the $L \lo 2$-space with respect to the distribution    $P \lo {X \hi {-(i,j)}}$ of $X \hi {-(i,j)}$. As shown in \citet{lee2013general}, the closure of the range of the regression operator is equal to the central subspace; that is,
\begin{align}\label{eq:regression operator vs central class}
\blue{\overline\ran ( R \lo {X \hi {-(i,j)} X \hi {(i,j)}} ) = \mathfrak S \lo {X \hi {(i,j)} | X \hi {-(i,j)}}}
\end{align}
 under the following assumption.
\begin{assumption}  \ \
\begin{enumerate}
\item $\sten H \lo X \hi {-(i,j)}$ is dense in $L \lo 2 (P \lo {X \hi {-(i,j)}} )$ modulo constants; that is, for any $f \in L \lo 2 (P \lo {X \hi {-(i,j)}} )$ and any $\epsilon > 0$, there is a $g \in \sten H \lo X \hi {-(i,j)}$ such that $\var [ f( X \hi {-(i,j)} )  - g( X \hi {-(i,j)} ) ] < \epsilon$;
\item $\mathfrak S \lo {X \hi {(i,j)} | X \hi {-(i,j)}}$ is a sufficient and complete.
\end{enumerate}
\end{assumption}
\blue{
The first condition   essentially requires the kernel $\ka \lo X \hi {-(i,j)}$ to be a universal kernel with respect to the $L \lo 2(P \lo {X \hi {-(i,j)}})$-norm. It means $\sten H \hi {-(i,j)}$ is rich enough to approximate any $L \lo 2(P \lo {X \hi {-(i,j)}})$-function arbitrarily closely. For example, it is satisfied by the Gaussian radial basis function kernel, but  not by the polynomial kernel.  For more information on universal kernels, see \citet*{sriperumbudur2011universality}. The completeness in the second condition means
\begin{align*}
E[ g (X \hi {-(i,j)}) | X \hi {(i,j)}] = 0 \ \mbox{almost surely} \ \Rightarrow \ g (X \hi {-(i,j)}) = 0 \ \mbox{almost surely}.
\end{align*}
This concept is defined in \citet*{lee2013general},
and is similar to the classical definition of completeness treating $X \hi {-(i,j)}$ as the parameter. \citet*{lee2013general} showed that completeness is a mild condition, and is satisfied by most nonparametric models.}

A basis of the central class $\mathfrak S \lo {X \hi {(i,j)} | X \hi {-(i,j)}}$ can be found by
solving  the generalized eigenvalue problem: for $k = 1, \ldots, d \lo {ij}$,
\begin{align}\label{eq:gep for gsir}
\begin{split}
\mbox{maximize} \quad \ali \langle f, \Sigma \lo {X \hi {-(i,j)} X \hi { (i,j)} } A \Sigma \lo {X \hi { (i,j)} X \hi {-  (i,j)} } f  \rangle \lo {-(i,j)}\\
\mbox{subject to} \quad \ali
\begin{cases}
\langle f \lo k,  \Sigma \lo {X \hi {-(i,j)} X \hi {-(i,j)} } f \lo k \rangle \lo {-(i,j)} = 1 \\
\langle f \lo k,  \Sigma \lo {X \hi {-(i,j)} X \hi {-(i,j)} } f \lo \ell \rangle \lo {-(i,j)}, \ \mbox{for } \ell=1, \ldots, k-1
\end{cases}
\end{split}
\end{align}
where $A: \sten H \lo X \hi {(i,j)} \to \sten H \lo X \hi {(i,j)}$ is any nonsingular and self adjoint operator, and $\langle \cdot, \cdot \rangle \lo {-(i,j)}$ is the inner product in $\sten H \lo X \hi {-(i,j)}$. That is, if $f \hi {ij} \lo 1, \ldots f \hi {ij} \lo {d \lo {ij}}$ are the first $d \lo {ij}$ eigenfunctions of this eigenvalue problem, then they span the central class. This type of \blue{estimate} of the central class is called generalized sliced inverse regression.
Convenient choices of $A$ are the identity mapping $I$ or the operator $\Sigma \lo {X \hi {(i,j)} X \hi {(i,j)}} \inv$. If we use the latter, then we need the following assumption.

\begin{assumption}\label{assumption:ran in ran 1} $\ran ( \Sigma \lo {X \hi {(i,j)} X \hi {-(i,j)} } ) \subseteq \ran ( \Sigma \lo {X \hi {(i,j)} X \hi {(i,j)} } )$.
\end{assumption}
\blue{This assumption has the similar interpretation as Assumption \ref{assumption:ran in ran}; see Section \rm S11 in the Supplementary Material.
At the population level, choosing $A$ to be $\Sigma \lo {X \hi {-(i,j)} X \hi {-(i,j)}} \inv$ achieves  better scaling because it down weights those components of the output of  $\Sigma \lo {X \hi {-(i,j)}X \hi {(i,j)}}$ with larger  variances. However, if the sample size is not sufficiently large, involving an estimate of $\Sigma \lo {X \hi {-(i,j)}X \hi {(i,j)}}\inv$ in the procedure could incur extra variations that  overwhelm the benefit brought by $\Sigma \lo {X \hi {-(i,j)}X \hi {(i,j)}}\inv $. In this case,  a nonrandom operator such as $A=I$ is preferable.}
In this paper we use  $A = \Sigma \lo { X \hi {(i,j)} X \hi {(i,j)} } \inv$. Let $U \hi {ij}$ denote the random vector
$
(  f \hi {ij} \lo 1 (X \hi {-(i,j)}) , \ldots f \hi {ij} \lo {d \lo {ij}}(X \hi {-(i,j)}) ).
$
The set of random vectors $\{ U \hi {ij}: (i,j) \in \Gamma \times \Gamma, i > j \}$ is the output for the nonlinear sufficient dimension reduction step.

\subsection{Step 2:Estimation of sufficient graphical model}

To estimate the edge set of the sufficient graphical model
we need to find a way to determine whether $X \hi i \indep X \hi j | U \hi {ij}$ is true. We use a linear operator introduced by \citet{fukumizu2008kernel} to perform this task, which is briefly described as follows.
Let $U$, $V$, $W$ be random vectors taking values in measurable spaces $(\Omega \lo U, \ca F \lo U)$, $(\Omega \lo V, \ca F \lo V)$, and $(\Omega \lo W, \ca F \lo W)$.
Let $\Omega \lo {UW} = \Omega \lo U \times \Omega \lo W$, $\Omega \lo {VW} = \Omega \lo V \times \Omega \lo W$, $\ca F \lo {UW}= \ca F \lo U \times \ca F \lo V$, and $\ca F \lo {VW } = \ca F \lo V \times \ca F \lo W$.
Let
\begin{align*}
\ka \lo {UW}: \Omega \lo {UW} \times \Omega \lo {UW} \to \real, \quad
\ka \lo {VW}: \Omega \lo {VW} \times \Omega \lo {VW} \to \real, \quad
\ka \lo W: \Omega \lo W \times \Omega \lo W \to \real
\end{align*}
be positive kernels. For example, for $(u \lo 1, w \lo 1), (u \lo 2, w \lo 2) \in \Omega \lo {UW} \times \Omega \lo {UW}$, $\ka \lo {UW}$ returns a real number denoted by $\ka \lo {UW}[(u \lo 1, w \lo 1), (u \lo 2, w \lo 2)]$.  Let $\sten H \lo {UW}$, $\sten H \lo {VW}$, and $\sten H \lo W$ be the centered reproducing kernel Hilbert space's generated by the kernels $\ka \lo {UW}$, $\ka \lo {VW}$, and $\ka \lo W$.
Define the covariance operators
\begin{align}\label{eq:4 cov operators}
\begin{split}
\ali \Sigma \lo {(UW)(VW)}: \sten H \lo {VW} \to \sten H \lo {UW}, \quad \Sigma \lo {(UW)W}: \sten H \lo W \to \sten H \lo {UW}, \\
\ali  \Sigma \lo {(VW)W}: \sten H \lo W \to \sten H \lo {VW}, \quad  \Sigma \lo {WW}: \sten H \lo W \to \sten H \lo W
\end{split}
\end{align}
as before.
The following definition is due to \citet{fukumizu2008kernel}. Since it plays a special role in this paper,   we give it a name -- ``conjoined conditional covariance operator'' that figuratively depicts its form.

\begin{definition}\label{definition:conjoined} Suppose
\begin{enumerate}
\item If $S$ is $W$, or $(U,W)$, or $(V, W)$, then $E [ \ka \lo S (S, S) ] < \infty$;
\item $\ran(\Sigma \lo {W (VW)} ) \subseteq \ran (\Sigma \lo {WW})$,   $\ran (\Sigma \lo {W (UW)} ) \subseteq \ran (\Sigma \lo {WW})$.
\end{enumerate}
Then the operator
$
\Sigma \lo {\ddot U \ddot V|W} = \Sigma \lo {(UW)(VW)} - \Sigma \lo {(UW)W} \Sigma \lo {WW} \inv \Sigma \lo {W(VW)}
$
is called the conjoined conditional covariance operator between $U$ and $V$ given $W$.
\end{definition}

The word ``conjoined'' describes the peculiar way in which $W$ appears in $\Sigma \lo {(UW)W}$ and $\Sigma \lo {W(VW)}$, which differs from an ordinary conditional covariance operator, where these operators are replaced by $\Sigma \lo {UW}$ and $\Sigma \lo {WV}$. The following proposition is due to \citet{fukumizu2008kernel}, a proof of a special case of which is given in \citet{fukumizu2004dimensionality}.

\begin{proposition}\label{theorem:conjoined equivalence} Suppose
\begin{enumerate}
\item $\sten H \lo {UW} \otimes \sten H \lo {VW}$ is probability determining;
\item for each $f \in \sten H \lo {UW}$, the function $ E[ f(U, W) | W=\cdot]$ belongs to $\sten H \lo W$;
\item for each $g \in \sten H \lo {VW}$, the function $E[ g(V, W) | W =\cdot ]$ belongs to $\sten H \lo W$;
\end{enumerate}
Then $\Sigma \lo {\ddot U \ddot V|W} = 0$ if and only if $U \indep V | W$.
\end{proposition}

\blue{The notion of probability determining in the context of reproducing kernel Hilbert space was defined in  \citet{fukumizu2004dimensionality}. For a generic random vector $X$, an reproducing kernel Hilbert space $\sten H\lo X$ based on a kernel $\ka \lo X$ is probability determining  if and only if the mapping
$P \mapsto E \lo P [\ka \lo X(\cdot, X)]$
is injective.
Intuitively, this requires the family of expectations $\{ E \lo P f(X): f \in \ca H \lo X \}$ to be rich enough to  identify  $P$. For example,  the Gaussian radial basis function is probability determining, but a polynomial kernel is not.} We apply the above proposition to $X \hi i, X \hi j, U \hi {ij}$ for each $(i,j) \in \Gamma \times \Gamma$, $i > j$. Let
\begin{align*}
\ka \lo {XU} \hi {i,ij}: (\Omega \lo {X \hi i} \times \Omega \lo {U \hi {ij}} ) \times (\Omega \lo {X \hi i} \times \Omega \lo {U \hi {ij}} )  \to \real
\end{align*}
be a positive definite kernel, and $\sten H \lo {XU} \hi {i,ij}$ the centered reproducing kernel Hilbert space generated by $\ka \lo {XU} \hi {i,ij}$. Similarly, let
$
\ka \lo {U} \hi {ij}: \Omega \lo {U \hi {ij}} \times \Omega \lo {U \hi {ij}} \to \real
$
 be a positive  kernel, and $\sten H \lo {U} \hi {ij}$ the centered reproducing kernel Hilbert space generated by $\ka \lo {U} \hi {ij}$.

\begin{assumption}\label{assumption:1 2 1 2 3} Conditions (1) and (2) of Definition \ref{definition:conjoined} and conditions (1), (2), and (3) of Proposition \ref{theorem:conjoined equivalence} are satisfied with $U$, $V$, and $W$ therein replaced by
$X \hi i$, $X \hi j$, and $U \hi {ij}$, respectively, for each $(i,j) \in \Gamma \times \Gamma$ and $i > j$.
\end{assumption}
Under this assumption, the conjoined conditional covariance operator $\Sigma \lo {\ddot X \hi i \ddot X \hi j | U \hi {ij}}$ is well defined and has the following property.

\begin{corollary}
Under Assumption \ref{assumption:1 2 1 2 3}, we have\kyongwon{
$(i,j) \notin \mathcal E \Leftrightarrow \Sigma \lo {\ddot X \hi i \ddot X \hi j | U \hi {ij} } = 0.$}
\end{corollary}

This corollary motivates us to estimate the graph by thresholding the norm of the estimated conjoined conditional covariance operator.

\section{Estimation: sample-level implementation}\label{section:estimation-sample}

\subsection{Implementation of step 1}\label{subsection:Implementation of stap 1}

Let $(X \lo 1, Y \lo 1), \ldots, (X \lo n, Y \lo n)$ be an i.i.d. sample of $(X,Y)$. At the sample level, the centered reproducing kernel Hilbert space $\sten H \lo X \hi {-(i,j)}$ is spanned by the functions
\begin{align}\label{eq:spanning - i j}
\{ \ka \lo X \hi {-(i,j)} ( \cdot, X \hi {-(i,j)} \lo a ) - E \lo n [\ka \lo X \hi {-(i,j)} ( \cdot, X \hi {-(i,j)})]: a = 1, \ldots, n \},
\end{align}
where $\ka \lo X \hi {-(i,j)} (\cdot, X \hi {-(i,j)} )$ stands for the function $u \mapsto  \ka \lo X \hi {-(i,j)} (u, X \hi {-(i,j)} )$, and
 $E \lo n [ \ka \lo X \hi {-(i,j)} (\cdot, X \hi {-(i,j)} )]$  the function $u \mapsto E \lo n [ \ka \lo X \hi {-(i,j)} (u, X \hi {-(i,j)} )]$.

We estimate the covariance operators   $\Sigma \lo {X \hi {-(i,j)} X \hi {(i,j)}}$ and $\Sigma \lo { X \hi {-(i,j)} X \hi {-(i,j)}}$ by
\begin{align*}
 \hat \Sigma \lo {X \hi {-(i,j)} X \hi {(i,j)} } =
\ali
E \lo n  \{[ \ka \lo X \hi {-(i,j)} ( \cdot, X  \hi {-(i,j)} )
-E \lo n  \ka \lo X \hi {-(i,j)} ( \cdot, X  \hi {-(i,j)} )] \\
\ali \otimes
[ \ka \lo X \hi {(i,j)} ( \cdot, X  \hi {(i,j)} )
-E \lo n  \ka \lo X \hi {(i,j)} ( \cdot, X  \hi { (i,j)} )] \} \\
 \hat \Sigma \lo { X \hi {-(i,j)} X \hi {-(i,j)}} = \ali
E \lo n \{ [ \ka \lo X \hi {-(i,j)} ( \cdot, X  \hi {-(i,j)} )
-E \lo n  \ka \lo X \hi {-(i,j)} ( \cdot, X  \hi {-(i,j)} )] \\
\ali \otimes
[ \ka \lo X \hi {-(i,j)} ( \cdot, X  \hi {-(i,j)} )
-E \lo n  \ka \lo X \hi {-(i,j)} ( \cdot, X  \hi {-(i,j)} )] \},
\end{align*}
respectively. We estimate $\Sigma \lo {X \hi {(i,j)} X \hi {(i,j)}} \inv $ by the Tychonoff-regularized inverse
$
 ( \hat \Sigma \lo {X \hi {(i,j)} X \hi { (i,j)} } + \epsilon \lo X \hi {(i,j)} I )\inv,
$
where $I: \sten H \lo X \hi {(i,j)} \to \sten H \lo X \hi {(i,j)}$ is the identity operator.
\blue{The regularized inverse is used to avoid over fitting. It plays the same role as ridge regression \citep{hoerl1970ridge} that alleviates over fitting by adding a multiple of the identity matrix to the sample covariance matrix before inverting it. }
%
%

At the sample level, the generalized eigenvalue problem (\ref{eq:gep for gsir}) takes the following form: at the $k$th iteration,
\begin{align}\label{eq:eigen problem for U}
\begin{split}
\mbox{maximize} \quad \ali \langle f, \hat \Sigma \lo {X \hi {-(i,j)} X \hi { (i,j)} }  ( \hat \Sigma \lo {X \hi {(i,j)} X \hi { (i,j)} } + \epsilon \lo X \hi {(i,j)} I )\inv   \hat \Sigma \lo {X \hi { (i,j)} X \hi {-  (i,j)} } f \rangle \lo {-(i,j)} \\
\mbox{subject to} \quad \ali
\begin{cases}
\langle f,  \hat \Sigma \lo {X \hi {-(i,j)} X \hi {-(i,j)} } f \rangle \lo {-(i,j)} = 1, \\
\langle f,  \hat \Sigma \lo {X \hi {-(i,j)} X \hi {-(i,j)} } f  \lo \ell \rangle \lo {-(i,j)} = 0, \quad \ell = 1, \ldots, k-1,
\end{cases}
\end{split}
\end{align}
where $f \lo 1, \ldots, f \lo {k-1}$ are the maximizers in the previous steps. The first $d \lo {ij}$ eigenfunctions are an estimate of a basis in the central class $\frak S \lo {X \hi {(i,j)} | X \hi {-(i,j)}}$.

Let $K \lo {X \hi {-(i,j)}}$ be the $n \times n$ matrix whose $(a,b)$th entry is $\ka \lo X \hi {-(i,j)} (X  \lo a \hi {-(i,j)}, X \lo b \hi {-(i,j)})$,  $Q = I \lo n - 1 \lo n 1 \lo n \trans / n$, and
$G \lo {X \hi {-(i,j)}} = Q K \lo {X \hi {-(i,j)}} Q$.
Let $a \hi 1, \ldots, a \hi {d \lo {ij}}$ be the first $d \lo {ij}$ eigenvectors of the matrix
\begin{align*}
\ali ( G \lo {X \hi {-(i,j)}}   + \epsilon \lo X \hi {-(i,j)} I \lo n )\inv G \lo {X \hi {-(i,j)}} G \lo {X \hi {(i,j)}} ( G \lo  {X \hi { (i,j)}}   + \epsilon \lo X \hi {(i,j)} I \lo n )\inv  G \lo  {X \hi {-  (i,j)}} ( G \lo {X \hi {-(i,j)}}   + \epsilon \lo X \hi {-(i,j)} I \lo n )\inv.
\end{align*}
Let
$ b \hi r =  ( G \lo {X \hi {-(i,j)}} + \epsilon \lo X \hi {-(i,j)} I \lo n ) \inv a \hi r$  for $ r = 1, \ldots, d \lo {ij}$.
As shown in Section S12.2,  the eigenfunctions $f \lo 1 \hi {ij}, \ldots, f \lo {d \lo {ij}} \hi {ij}$ are calculated by
\begin{align*}
f \lo r \hi {ij} = \sum \lo {a=1} \hi n b \hi r \lo a \{ \ka \lo X \hi {-(i,j)} ( \cdot, X \hi {-(i,j)} \lo a ) - E \lo n [\ka \lo X \hi {-(i,j)} ( \cdot, X \hi {-(i,j)})]\}.
\end{align*}
The statistics $\hat U \hi {ij} \lo a  = ( f \lo 1 \hi {ij} (X \lo a \hi {-(i,j)}) , \ldots, f \lo {d \lo {ij}} \hi {ij} (X \lo a \hi {-(i,j)}))$, $a = 1, \ldots, n$, will be used as the input for the second step.

\subsection{Implementation of step 2}

This step  consists of estimating the conjoined conditional covariance operator's for each $(i,j)$ and thresholding their norms. At the sample level,  the centered reproducing kernel Hilbert space's generated by the kernels $\ka \lo {XU} \hi {i,ij}$, $\ka \lo {XU} \hi {j,ij}$, and  $\ka \lo U \hi {ij}$ are
\begin{align*}
\sten H \lo {XU} \hi {\,i,ij}= \ali  \spn \{\ka \lo {XU} \hi {i,ij} ( \cdot, (X \lo a \hi i, U \lo a  \hi {ij})) - E \lo n [ \ka \lo {XU} \hi {i,ij} ( \cdot, (X \hi i, U \hi {ij})) ]: a = 1, \ldots, n  \}, \\
\sten H \lo {XU} \hi {\,j,ij}= \ali  \spn \{\ka \lo {XU} \hi {j,ij} ( \cdot, (X \lo a \hi j, U \lo a \hi {ij})) - E \lo n [ \ka \lo {XU} \hi {j,ij} ( \cdot, (X \hi j, U \hi {ij})) ]: a = 1, \ldots, n  \}, \\
\sten H \lo {U} \hi {\,ij}= \ali  \spn \{\ka \lo {U} \hi {ij} ( \cdot,   U \lo a \hi {ij}) - E \lo n [ \ka \lo {U} \hi {ij} ( \cdot, U \hi {ij}) ]: a = 1, \ldots, n  \},
\end{align*}
where, for example, $\ka \lo {XU} \hi {i,ij} ( \cdot, (X \lo a \hi i, U \lo a  \hi {ij}))$ denotes  the function
\begin{align*}
\Omega \lo {X \hi i} \times \Omega \lo {U \hi {ij}} \to \real, \quad (x \hi i, u \hi {ij} ) \mapsto \ka \lo {XU} \hi {i,ij} ( (x \hi i, u \hi {ij} ), (X \lo a \hi i, U \lo a  \hi {ij}))
\end{align*}
and $E \lo n [ \ka \lo {XU} \hi {i,ij} ( \cdot, (X \hi i, U \hi {ij})) ]$ denotes the function
\begin{align*}
\Omega \lo {X \hi i} \times \Omega \lo {U \hi {ij}} \to \real, \quad (x \hi i, u \hi {ij} ) \mapsto E \lo n [ \ka \lo {XU} \hi {i,ij} ( (x \hi i, u \hi {ij} ), (X   \hi i, U  \hi {ij}))].
\end{align*}

We estimate the covariance  operators
$\Sigma \lo {(X \hi i U \hi {ij})( X \hi i U \hi {ij})}$,  $\Sigma \lo {(X \hi i U \hi {ij})U \hi {ij}}$,   $\Sigma \lo {X \hi j (X \hi jU \hi {ij})  }$, and $\Sigma \lo {U \hi {ij} U \hi {ij}}$ by
\begin{align}\label{eq:estimated operators}
\begin{split}
\hat  \Sigma \lo {(X \hi i U \hi {ij}) (X \hi j U \hi {ij})} = \ali  E \lo n  \{  [  \ka \lo {XU} \hi {i,ij} ( \cdot, ( X \hi i, U \hi {ij}))-  E \lo n  \ka \lo {XU} \hi {i,ij} ( \cdot, ( X \hi i, U \hi {ij}))  ] \\
\ali \hspace{.2in} \otimes [  \ka \lo {XU} \hi {j,ij} ( \cdot, ( X \hi j, U \hi {ij}))-  E \lo n  \ka \lo {XU} \hi {j,ij} ( \cdot, ( X \hi j, U \hi {ij}))  ] \}  \\
\hat  \Sigma \lo {(X \hi i U \hi {ij}) U \hi {ij}} = \ali  E \lo n  \{  [  \ka \lo {XU} \hi {i,ij} ( \cdot, ( X \hi i, U \hi {ij}))-  E \lo n  \ka \lo {XU} \hi {i,ij} ( \cdot, ( X \hi i, U \hi {ij}))  ] \\
\ali \hspace{.2in} \otimes [  \ka \lo {U} \hi {ij} ( \cdot,  U \hi {ij})-  E \lo n  \ka \lo {U} \hi {ij} ( \cdot, U \hi {ij}) ] \}  \\
\hat   \Sigma \lo {U \hi {ij}(X \hi j  U \hi {ij})  } = \ali  E \lo n  \{  [  \ka \lo {U} \hi {ij} ( \cdot,  U \hi {ij})-  E \lo n  \ka \lo {U} \hi {ij} ( \cdot, U \hi {ij}) ]  \\
\ali \hspace{.2in} \otimes [  \ka \lo {XU} \hi {j,ij} ( \cdot, ( X \hi j, U \hi {ij}))-  E \lo n  \ka \lo {XU} \hi {j,ij} ( \cdot, ( X \hi j, U \hi {ij}))  ] \}   \\
\hat   \Sigma \lo {U \hi {ij} U \hi {ij}}  = \ali E \lo n  \{ [  \ka \lo {U} \hi {ij} ( \cdot,  U \hi {ij})-  E \lo n  \ka \lo {U} \hi {ij} ( \cdot, U \hi {ij}) ]  \\
\ali \hspace{.2in} \otimes [  \ka \lo {U} \hi {ij} ( \cdot,  U \hi {ij})-  E \lo n  \ka \lo {U} \hi {ij} ( \cdot, U \hi {ij}) ]  \},
\end{split}
\end{align}
respectively. We then estimate the conjoined conditional covariance operator by
\begin{align*}
\hat \Sigma \lo {\ddot X \hi i \ddot X \hi j | U \hi {ij}}=
\hat  \Sigma \lo {(X \hi i U \hi {ij}) (X \hi j U \hi {ij})} -
\hat  \Sigma \lo {(X \hi i U \hi {ij}) U \hi {ij}}
(\hat   \Sigma \lo {U \hi {ij} U \hi {ij}}  + \epsilon \lo U \hi {(i,j)} I ) \inv
\hat   \Sigma \lo {U \hi {ij}(X \hi j  U \hi {ij})  },
\end{align*}
where, again,  we have used Tychonoff regularization to estimate the inverted covariance operator $ \Sigma \lo {U \hi {ij} U \hi {ij}}$.
Let $K \lo {U \hi {ij}}$, $K  \lo {X \hi i U \hi {ij}}$, and $K  \lo {X \hi j U \hi {ij}}$ be the Gram matrices
\begin{align*}
K \lo {U \hi {ij}}= \ali \{ \ka \lo U \hi {ij} (U \lo a \hi {ij}, U \lo b \hi {ij}) \} \lo {a, b = 1} \hi n, \\
K  \lo {X \hi i U \hi {ij}}= \ali \{\ka \lo {XU} \hi {i, ij} ((X \hi i \lo a, U \lo a \hi {ij}), (X \hi i \lo b, U \lo b \hi {ij})) \} \lo{a, b =1} \hi n, \\
K  \lo {X \hi j U \hi {ij}} = \ali \{ \ka \lo {XU} \hi {j, ij} ((X \hi j \lo a, U \lo a \hi {ij}), (X \hi j \lo b, U \lo b \hi {ij})) \} \lo {a,b=1} \hi n,
\end{align*}
and  $G \lo {X \hi i U \hi {ij}}$,
$G \lo {X \hi j U\hi {ij}}$, and
$G \lo {U \hi {ij}}$   their centered versions
\begin{align*}
G \lo {X \hi i U \hi {ij}} = Q K \lo {X \hi i U \hi {ij}} Q, \quad
G \lo {X \hi j U\hi {ij}} = Q K  \lo {X \hi jU \hi {ij}} Q, \quad
G \lo {U \hi {ij}} = Q K  \lo {U \hi {ij}} Q.
\end{align*}
As shown in Section \rm S12 in the Supplementary Material,
\begin{align*}
 \| \hat \Sigma \lo {\ddot X \hi i \ddot X \hi j| U \hi {ij}} \| \looo {hs}
=  \left \|G \lo {X \hi i U \hi {ij}} \hi {1/2} G \lo {X \hi j U \hi {ij}} \hi {1/2} - G \lo {X \hi i U \hi {ij}} \hi {1/2} G \lo {U \hi {ij}} (  G \lo {U \hi {ij}} + \epsilon \lo U \hi {(i,j)} Q ) \hi \dagger G \lo {X \hi {j}U\hi {ij}} \hi {1/2}   \right\| \looo f,
\end{align*}
where $\| \cdot \| \looo f$ is the Frobenius norm.
Estimation of the edge set is then based on thresholding this norm; that is,
\begin{align*}
\hat {\ca E} = \{ (i,j) \in \Gamma  \times \Gamma: \, i > j, \  \| \hat \Sigma \lo {\ddot X \hi i \ddot X \hi j| U \hi {ij}} \| \looo {hs} > \rho \lo n \}
\end{align*}
for some chosen $\rho \lo n > 0$.

\subsection{Tuning} \label{sec;tuning}
We have \blue{three} types of tuning constants: those for the kernels, those for Tychonoff regularization, \blue{and the threshold $\rho \lo n$}. For the Tychonoff regularization, we have   $\epsilon \lo X \hi {(i,j)}$ and  $\epsilon \lo X \hi {-(i,j)}$ for step 1, and $\epsilon \lo U \hi {(i,j)}$ for step 2. In this paper we  use the Gaussian radial basis function as the  kernel:
\begin{align}\label{eq:rbf}
\ka (u,v) = \exp ( - \gamma \| u - v  \| \hi 2 ).
\end{align}
For each $(i,j)$, we  have five $\gamma$'s to determine: $\gamma \lo X \hi {(i,j)}$ for the kernel $\ka \lo X \hi {(i,j)}$, $\gamma \lo X \hi {-(i,j)}$ for $\ka \lo X \hi {-(i,j)}$, $\gamma \lo {XU} \hi {i,ij}$ for $\ka \lo {XU} \hi {i,ij}$,  $\gamma \lo {XU} \hi {j,ij}$ for  $\ka \lo {XU} \hi {j,ij}$, and $\gamma \lo U \hi {ij}$ for $\ka \lo U \hi {ij}$, which are chosen by the following formula (see, for example, \citet{li2018sufficient})
\begin{align}\label{eq:tuning gamma}
1/ \sqrt \gamma = {n \choose 2} \inv \sum \lo {a < b} \| s \lo a - s \lo b \|,
\end{align}
where  $s \lo 1, \ldots, s \lo n$ are the sample of random vectors corresponding to the mentioned five kernels. For example, for the kernel $\ka \lo {XU}  \hi {j, ij}$, $s \lo a =  (X  \lo a \hi j, U \lo a \hi {ij})$.
For the tuning parameters in Tychonoff regularization, we use the following generalized cross validation scheme (GCV; see \cite{golub-heath-wahba-1979}):
\begin{align}\label{eq:GCV epsilon}
\text{GCV}(\epsilon)= \argmin\lo{\epsilon} \sum\lo{i<j}\frac{ \Vert  G\lo 1-G\lo 2\trans [ G\lo 2+\epsilon \hspace{0.5mm} \lambda\lo{\max}(G\lo 2)]\hi{-1}G\lo 1
\Vert\looo{F}}{\frac{1}{n}\text{tr}\{I\lo n-G\lo 2\trans [ G\lo 2+\epsilon \hspace{0.5mm} \lambda\lo{\max}(G\lo 2) ]\hi{-1}\}},
\end{align}
where $G \lo 1, G \lo 2 \in \real \hi {n \times n}$ are positive semidefinite matrices, and $\lambda \lo \max (G \lo 2)$ is the largest eigenvalue of $G \lo 2$. The matrices $G \lo 1$ and $G \lo 2$ are the following matrices for the three tuning parameters:
\begin{enumerate}
\item $G \lo 1 = G \lo {X \hi {-(i,j)}}$, $G \lo 2 = G \lo {X \hi {(i,j)}}$ for $\epsilon \lo X \hi {(i,j)}$,
\item $G \lo 1 = G \lo {X \hi {(i,j)}}$, $G \lo 2 = G \lo {X \hi {-(i,j)}}$ for $\epsilon \lo X \hi {-(i,j)}$,
\item $G \lo 1 = G \lo {X \hi {(i,j)}}$, $G \lo 2 = G \lo {U \hi {ij}}$ for $\epsilon \lo U \hi {(i,j)}$,
\end{enumerate}
We minimize (\ref{eq:GCV epsilon}) over a grid to choose $\epsilon$, as detailed in Section \ref{sec;simulation}.

\blue{We also use
generalized cross validation to determine the thresholding parameter $\rho \lo n$. Let $\hat{\ca E}(\rho)$ be the estimated edge set using a threshold $\rho$, and, for each $i \in \Gamma$,  let  $C \hi i (\rho)=\{ X \hi j: \, j \in \Gamma,  \, (i,j) \in \hat{\ca E}(\rho) \}$  be the subset of components of $X$ at the neighborhood of the node   $i$ in the graph $(\Gamma, \hat E ( \rho))$. The basic idea is to apply  the generalized cross validation  to the  regression of the feature of $X \hi i$  on the feature of  $C \hi i (\rho)$. The generalized cross validation  for this regression takes the form
\begin{align}\label{eq:GCV for rho}
\text{GCV} (\rho) = \sum\lo{i=1}\hi p\frac{ \Vert  G\lo{X\hi i}-G\lo{C \hi i (\rho)}\trans [ G\lo{C \hi i (\rho)}+\epsilon \hspace{0.5mm} \lambda\lo{\max}(G\lo{C \hi i (\rho)})I \lo n ]\hi{-1}G\lo{X\hi i}
\Vert\looo{F}}{\frac{1}{n}\text{tr}\{I\lo n-G\lo{C \hi i (\rho)}\trans [ G\lo{C \hi i (\rho)}+\epsilon \hspace{0.5mm} \lambda\lo{\max}(G\lo {C \hi i (\rho) } ) I \lo n ]\hi{-1}\}},
\end{align}
where $G \lo {C \hi i (\rho)}= Q K \lo {C \hi i (\rho)} Q$,  and $K \lo {C \hi i (\rho)}$ is the $n \times n$ kernel matrix for the sample of $C \hi i (\rho)$.
We minimize $\text{GCV}(\rho)$ over the grid  $\rho\in \{\ell \times 10\hi{-2}:  \ell=2, \dots, 7\}$ to determine the optimal threshold $\rho \lo n$.

Regarding the selection of the dimension of $U \hi {ij}$,   to  our knowledge there has been no systematic procedure available to determine the dimension of the central class for nonlinear sufficient dimension reduction. While some recently developed methods for order determination for linear sufficient dimension reduction, such as the ladle estimate  and predictor augmentation estimator \citep{luo2016combining,luo2020order}, may be generalizable  to the nonlinear sufficient dimension reduction setting, we will leave this topic to future research. Our experiences and intuitions indicate that a small dimension, such as 1 or 2, for the central class would be sufficient in most cases. For example, in the classical nonparametric regression problems  $Y = f(X) + \epsilon$ with $X \indep \epsilon$, the dimension of the central class is by definition equal to 1.
}


\section{Asymptotic theory}\label{section:asymptotics}

In this section we develop the consistency and convergence rates of our estimate and related operators. The challenge of this analysis is that our  procedure involves two steps:  we first  extract the sufficient predictor  using one set of kernels, and then substitute it into another set of kernels to get the final result. Thus we need to understand how the error propagates  from the first step to the second. \red{We also develop the asymptotic theory allowing $p$ to go to infinity with $n$, which is presented in the Supplementary Material.}

\subsection{Overview}\label{subsection:asymptotics}

Our goal is to derive the convergence rate of
\begin{align*}
\left| \| \hat \Sigma \lo {\ddot X \hi i \ddot X \hi j | \hat U \hi {ij}} \| \looo {hs} - \|  \Sigma \lo {\ddot X \hi i \ddot X \hi j |  U \hi {ij}} \| \looo {hs} \right|,
\end{align*}
as $\| \hat \Sigma \lo {\ddot X \hi i \ddot X \hi j | \hat U \hi {ij}} \| \looo {hs}$ is the quantity we threshold to determine the edge set.
By the triangular inequality,
\begin{align*}
\ali \left| \| \hat \Sigma \lo {\ddot X \hi i \ddot X \hi j | \hat U \hi {ij}} \| \looo {hs} - \|  \Sigma \lo {\ddot X \hi i \ddot X \hi j |  U \hi {ij}} \| \looo {hs} \right|
\le
 \| \hat \Sigma \lo {\ddot X \hi i \ddot X \hi j | \hat U \hi {ij}} -  \Sigma \lo {\ddot X \hi i \ddot X \hi j |  U \hi {ij}} \| \looo {hs}  \\
\ali \hspace{1in}\le
\| \hat \Sigma \lo {\ddot X \hi i \ddot X \hi j | \hat U \hi {ij}}-  \hat \Sigma \lo {\ddot X \hi i \ddot X \hi j |  U \hi {ij}} \| \looo {hs} + \| \hat \Sigma \lo {\ddot X \hi i \ddot X \hi j |  U \hi {ij}} -   \Sigma \lo {\ddot X \hi i \ddot X \hi j |  U \hi {ij}} \| \looo {hs}.
\end{align*}
So we need to derive the  convergence rates of the following quantities:
\begin{align}\label{eq:three convergence}
\begin{split}
\ali  \mbox{(i)} \quad  \| \hat U \hi {ij} - U \hi {ij} \| \lo {[\stens H \hi {-(i,j)} (X)] \hi {d \lo {ij}}}, \\
\ali \mbox{(ii)}  \quad   \| \hat \Sigma \lo {\ddot X \hi i \ddot X \hi j | \hat U \hi {ij}}-  \hat \Sigma \lo {\ddot X \hi i \ddot X \hi j |  U \hi {ij}} \| \looo {hs}, \\
\ali \mbox{(iii)} \quad   \| \hat \Sigma \lo {\ddot X \hi i \ddot X \hi j |  U \hi {ij}} -   \Sigma \lo {\ddot X \hi i \ddot X \hi j |  U \hi {ij}} \| \looo {hs},
\end{split}
\end{align}
where, to avoid overly crowded subscripts, we have used $\sten H \hi {-(i,j)} (X)$ to denote $\sten H \hi {-(i,j)} \lo X$ when it occurs as a subscript.
The first and third convergence rates can be derived using the asymptotic tools for linear operators developed in \citet{fukumizu2007statistical}, \citet{li2017nonlinear}, \citet{lee2016}, and \citet{solea2020copula}. The second convergence rate is, however,  a new problem,  and it will also be useful in similar settings that require   constructing estimators based on  predictors extracted by sufficient dimension reduction. In some sense, this is akin to the post dimension reduction problem considered in \citet{kim2019post}.

\def\O{\dot O}
\def\o{\dot o}
\def\ddO{\ddot O}
\def\ddo{\ddot o}

In the following,  if $\{a \lo n \}$ and $\{  b \lo n \}$ are sequences of positive numbers, then we  write $a \lo n \prec b \lo n$ if $a \lo n / b \lo n \to 0$. We  write $a \lo n \asymp b \lo n$ if $0< \liminf \lo n (b \lo n / a \lo n) \le  \limsup \lo n (b \lo n / a \lo n)  < \infty$. We write $b \lo n \preceq a \lo n$ if either $b \lo n \prec a \lo n$ or $b \lo n \asymp a \lo n$. Because $(i,j)$ is fixed in the asymptotic development, and also to emphasize the dependence on $n$, in the rest of this section we denote $\epsilon \lo X \hi {(i,j)}$, $\epsilon \lo X \hi {-(i,j)}$, and $\epsilon \lo U \hi {(i,j)}$ by $\epsilon \lo n$, $\eta \lo n$, and $\delta \lo n$, respectively.


\subsection{Transparent kernel}

We first develop what we call the ``transparent kernel'' that passes information  from step 1 to step 2 efficiently. Let $\Omega$ be a nonempty set, and $\ka : \Omega \times \Omega \to \real$ a positive kernel.

\begin{definition} We say that $\ka$ is a transparent kernel if, for each $t \in \Omega$,  the function $s \mapsto \ka (s,t)$ is twice differentiable and
\begin{enumerate}
\item  $\partial \ka (s,t)/ \partial s | \lo {s=t} = 0$;
\item the matrix $H(s,t) = \partial \hi 2 \ka (s,t) / \partial s \partial s \trans$ has a bounded operator norm; that is, there exist $-\infty < C \lo 1 \le  C \lo 2 < \infty$ such that
\begin{align*}
C \lo 1 \le \lambda \lo \min (H(s,t)) \le \lambda \lo \max (H(s,t)) < C \lo 2
\end{align*}
 for all $(s,t) \in \Omega \times \Omega$, where $\lambda \lo \min(\cdot)$ and $\lambda \lo \max (\cdot)$ indicate the largest and smallest eigenvalues.
\end{enumerate}
\end{definition}

\noindent
\blue{For example, the Gaussian radial basis function kernel is   transparent, but the exponential kernel
$
        \ka(u,v) = \tau \hi 2 \exp(-\gamma \Vert u-v \Vert )
$  is not.}
This condition implies a type of Lipschitz continuity in a setting that involves  two reproducing kernels $\ka \lo 0$ and $\ka \lo 1$, where  the argument of $\ka \lo 1$ is the evaluation of a member of the reproducing kernel Hilbert space generated by $\ka \lo 0$.

\begin{theorem}\label{lemma:change norms}
Suppose $\sten H \lo 0$ is the reproducing kernel Hilbert space generated by $\ka \lo 0$, $\sten H \lo 0 \hi {\, d}$ is the $d$-fold Cartesian product of $\sten H \lo 0$ with inner product defined by
\begin{align*}
\langle U, V \rangle \lo {\stens H \lo 0 \hi {\ d}} = \langle u \lo 1, v \lo 1 \rangle \lo {\stens H \lo 0} + \cdots + \langle u \lo d, v \lo d \rangle \lo {\stens H \lo 0}
\end{align*}
where $U = (u \lo 1, \ldots, u \lo d)$ and $V = (v \lo 1, \ldots, v \lo d)$ are members of $\sten H \lo 0 \hi {\, d}$,
$\sten H \lo 1$ is the reproducing kernel Hilbert space generated by $\ka \lo 1$. Then:
\begin{enumerate}
\item[(i)] for any   $U, V \in \sten H \lo 0 \hi {\, d}, \ a \in \Omega$, we have
\begin{align*}
\| U (a)- V(a) \| \lo {\real \hi {\, d}}
\le [ \ka \lo 0(a, a) ] \hi {1/2} \
\| U- V  \| \lo {\stens H \lo 0\hi {\ d}};
\end{align*}
\item[(ii)]
if  $\ka \lo 1(s,t) $ is a transparent kernel,
then there exists a   $C> 0$  such that, for each $U, V \in \sten H \lo 0 \hi {\, d}$ and $a \in \Omega$,
\begin{align*}
\ali \|  \ka \lo 1  ( \cdot, U ( a) ) - \ka \lo 1  ( \cdot, V ( a) ) \| \lo {\stens H \lo 1}    \le C \, [\ka \lo 0  (a , a)]\hi {1/2} \, \| U - V \| \lo {\stens H \lo 0 \hi {\,\, d}}.
\end{align*}
\end{enumerate}
\end{theorem}


A direct consequence of this theorem is that, if $\hat U$ is an estimate of some $U $, a member of $\sten H \lo 0 \hi d$,  with $\| \hat U - U \| \lo {{\stens H \lo 0 } \hi d } = O \lo P ( b \lo n)$ for some $0  < b \lo n \to 0$,  $\hat \Sigma (\hat U)$ is a linear operator estimated from the sample  $\hat U \lo 1, \ldots, \hat U \lo n$ (and perhaps some other random vectors), and $\hat \Sigma (U)$ is a linear operator estimated from the sample  $U \lo 1, \ldots, U \lo n$,   then,
 \begin{align}\label{eq:transparent}
 \| \hat \Sigma ( \hat U) - \hat \Sigma (U) \| \looo {hs} = O \lo P ( b \lo n).
\end{align}
This result is somewhat surprising, because sample estimates  such as $ \hat \Sigma ( \hat U) $ can be viewed as $E \lo n {\mathbb G} ( X, \hat U )$, where  $\hat U$ is an estimate of a function $U$ in a functional space with norm $\| \cdot \|$ and ${\mathbb G}$ is an operator-valued function. If $\| \hat U - U \| = O \lo P (b \lo n)$ for some $b \lo n \to 0$, then it is not necessarily true that
\begin{align*}
\| E \lo n {\mathbb G} ( X, \hat U) -  E \lo n {\mathbb G} ( X,  U) \| = O \lo P (b \lo n),
\end{align*}
particularly when $U$ is an infinite dimensional object. Yet relation (\ref{eq:transparent}) states exactly this. The reason behind this  is that the reproducing kernel property separates the function $\hat U $ and its argument $X \lo a$ (i.e. $\hat U (x) = \langle \hat U, \ka (\cdot, x) \rangle$), which implies a type of uniformity among $\hat U (X \lo 1), \ldots, \hat U (X \lo n)$. This point will be made clear in the proof in the Supplementary Material.
Statement (\ref{eq:transparent}) is made precise by the next theorem.

\begin{theorem}\label{theorem:operator UU} Suppose conditions (1) and (2) of Definition \ref{definition:conjoined} are satisfied with $U$, $V$, $W$ therein replaced by $X \hi i$, $X \hi j$, and $U \hi {ij}$. Suppose, furthermore:
\begin{enumerate}
\item[(a)] $\ka \lo U \hi {ij}$,  $\ka \lo {XU}  \hi {i,ij}$, and  $\ka \lo {XU}  \hi {j,ij}$ are transparent kernels;
\item[(b)] $ \| \hat U \hi {ij} - U \hi {ij} \| \lo {[\stens H \hi {\ -(i,j)} (X) ] \hi {d \lo {ij}}} = O \lo P ( b \lo n )$ for some $0 < b \lo n \to 0$.
\end{enumerate}
Then
\begin{enumerate}
\item[(i)] $\| \hat \Sigma \lo {  \hat U \hi {ij} \hat U \hi {ij} } -   \hat    \Sigma \lo {    U \hi {ij}   U \hi {ij} } \| \looo {hs}=O \lo P ( b \lo n  )$;
\item[(ii)]  $ \| \hat \Sigma \lo { (X \hi {i} \hat U \hi {ij}) \hat U \hi {ij} } -   \hat    \Sigma \lo {   (X \hi {i} U \hi {ij})    U \hi {ij} } \| \looo {hs}=O \lo P ( b \lo n )$;
\item[(iii)]  $ \| \hat \Sigma \lo {  (X \hi {i} \hat U \hi {ij})  (X \hi {j} \hat U \hi {ij}) } -   \hat    \Sigma \lo {   (X \hi {i} U \hi {ij})  (X \hi {j} U \hi {ij})  } \| \looo {hs}=O \lo P ( b \lo n )$.
\end{enumerate}
\end{theorem}

\medskip

Using  Theorem \ref{theorem:operator UU} we can derive the convergence rate of $\| \hat \Sigma \lo {\ddot X \hi i \ddot X \hi j | \hat U \hi {ij}}-  \hat \Sigma \lo {\ddot X \hi i \ddot X \hi j |  U \hi {ij}} \| \looo {hs}$.

\medskip

\begin{theorem}\label{theorem:ccco approximation 1} Suppose conditions in Theorem \ref{theorem:operator UU} are satisfied and, furthermore,
\begin{enumerate}
\item[(a)]
$\Sigma \lo {U \hi {ij}U \hi {ij}} \inv \Sigma \lo {U \hi {ij}(X \hi i U \hi {ij})}$ and $\Sigma \lo {U \hi {ij}U \hi {ij}} \inv \Sigma \lo {U \hi {ij}(X \hi j U \hi {ij})}$
are bounded linear operators;
\item[(b)] $b \lo n \preceq \delta \lo n \prec 1$.
\end{enumerate}
 Then
$
\| \hat \Sigma \lo {\ddot X \hi i \ddot X \hi j | \hat U \hi {ij}}-  \hat \Sigma \lo {\ddot X \hi i \ddot X \hi j |  U \hi {ij}} \| \looo {hs} = O \lo P ( b \lo n  ).
$
\end{theorem}

\medskip

Note that, unlike in Theorem \ref{theorem:operator UU}, where our assumptions imply
\begin{align*}
\Sigma \lo {X \hi {-(i,j)} X \hi {-(i,j)}}\inv \Sigma \lo {X \hi {-(i,j)} X \hi {(i,j)}}
\end{align*}
 is a finite-rank operator, here, we do not assume
$\Sigma \lo {U \hi {ij}(U \hi {ij})}\inv \Sigma \lo {U \hi {ij}(X \hi j U \hi {ij})}$ to be a finite-rank (or even Hilbert-Schmidt) operator; instead, we assume it to be a bounded operator.
This is because  $(X \hi j,  U \hi {ij})$ contains $U \hi {ij}$, which makes it unreasonable to assume $\Sigma \lo {U \hi {ij}\red{U \hi {ij}}}\inv \Sigma \lo {U \hi {ij}(X \hi j U \hi {ij})}$ to be finite-rank or Hilbert Schmidt. For example, when $X \hi j$ is a constant, $\Sigma \lo {U \hi {ij}(X \hi j U \hi {ij})}$ is the same as $\Sigma \lo {U \hi {ij}  U \hi {ij}}$ and  $\Sigma \lo {U \hi {ij} U \hi {ij}} \inv \Sigma \lo {U \hi {ij} U \hi {ij}}$ is not a Hilbert Schmidt operator, though it is  bounded.
Theorem \ref{theorem:ccco approximation 1} shows that convergence rate of (ii) in (\ref{eq:three convergence}) is the same as the convergence rate of (i) in (\ref{eq:three convergence}); it now remains to derive the convergence rate of (i) and (iii).

\subsection{Convergence rates of (i) and (iii) in (\ref{eq:three convergence})}

We first present the convergence rate of $\hat U \hi {ij}$ to $U \hi {ij}$. The proof is similar to that of Theorem 5 of \citet{li2017nonlinear} but with two differences. First, \citet{li2017nonlinear} took $A$ in (\ref{eq:gep for gsir}) to be $I$, whereas we take it to be $\Sigma \lo {YY}$. In particular, the generalized sliced inverse regression in \citet{li2017nonlinear} only has one tuning parameter $\eta \lo n$, but we have two tuning parameters $\eta \lo n$ and $\epsilon \lo n$. Second, \citet{li2017nonlinear} defined (in the current notation) $f \lo r \hi {ij}$ to be the eigenfunctions of
\begin{align*}
\Sigma \lo {X \hi {-(i,j)}X \hi {-(i,j)}}\inv \Sigma \lo {X \hi {-(i,j)}X \hi {(i,j)}}\Sigma \lo {X \hi {(i,j)}X \hi {(i,j)}} \inv \Sigma \lo {X \hi {(i,j)}X \hi {-(i,j)}}\Sigma \lo {X \hi {-(i,j)}X \hi {-(i,j)}}\inv,
\end{align*}
which is different from the generalized eigenvalue problem (\ref{eq:gep for gsir}).
For these reasons  we need to re-derive the convergence rate of $\hat U \hi {ij}$.

\medskip

\begin{theorem}\label{theorem:hat U rate} Suppose
\begin{enumerate}
\item[(a)] Assumption \ref{assumption:finite expectation of kernel} is satisfied; 
\item[(b)] $\Sigma \lo {X \hi {-(i,j)} X \hi {(i,j)}}$ is a finite-rank operator with
\begin{align*}
\ali \ran ( \Sigma \lo {X \hi {-(i,j)} X \hi {(i,j)} } ) \subseteq \ran ( \Sigma \lo {X \hi {-(i,j)} X \hi {-(i,j)} } \hi {\red{2}}), \\
\ali  \ran ( \Sigma \lo {X \hi {(i,j)} X \hi {-(i,j)} } ) \subseteq \ran ( \Sigma \lo {X \hi {(i,j)} X \hi {(i,j)} });
\end{align*}
\item[(c)] $n \hi {-1/2} \prec \eta \lo n \prec 1$, $n \hi {-1/2} \prec \epsilon  \lo n \prec 1$;
\item[(d)] for each $r = 1, \ldots, d \lo {ij}$, $\lambda \hi {ij} \lo 1 > \cdots > \lambda \hi {ij} \lo {d \lo {ij}}$.
\end{enumerate}
Then,
$
\| \hat U \hi {ij} - U \hi {ij} \| \lo {[\stens H \hi {-(i,j)} (X) ] \hi {d \lo {ij}}}=  O \lo P (
 \eta \lo n \hi {-3/2} \epsilon \lo n \hi {-1}  n \hi {-1} + \eta \lo n \hi {-1} n \hi {-1/2} +  \red{\eta \lo n} + \epsilon \lo n  ).
$
\end{theorem}

\medskip

An immediate consequence is that, under the transparent kernel assumption, the  $b \lo n$ in Theorem \ref{theorem:ccco approximation 1} is the same as this rate. We next derive the convergence rate in (iii) of (\ref{eq:three convergence}). This rate depends on the tuning parameter $\delta \lo n$ in the estimate of conjoined conditional covariance operator, and it reaches $b \lo n$ for the optimal choice of $\delta \lo n$.

\begin{theorem}\label{theorem:ccco approximation 2} Suppose conditions (1) and (2) of Definition \ref{definition:conjoined} are satisfied with $U$, $V$, $W$ therein replaced by $X \hi i$, $X \hi j$, and $U \hi {ij}$. Suppose, furthermore,
\begin{enumerate}
\item[(a)]
$
\Sigma \lo {U \hi {ij}U \hi {ij}} \inv \Sigma \lo {U \hi {ij}(X \hi i U \hi {ij})}$ and $ \Sigma \lo {U \hi {ij}U \hi {ij}} \inv \Sigma \lo {U \hi {ij}(X \hi j U \hi {ij})}
$
are bounded linear operators;
\item[(b)] $b \lo n \preceq \delta \lo n  \prec  1$.
\end{enumerate}
Then
$
\| \hat \Sigma \lo {\ddot X \hi i \ddot X \hi j |  U \hi {ij}}-  \Sigma \lo {\ddot X \hi i \ddot X \hi j |  U \hi {ij}} \| \looo {hs} = O \lo P (\delta \lo n).
$ Consequently, if $\delta \lo n \asymp b \lo n$, then
\begin{align*}
\| \hat \Sigma \lo {\ddot X \hi i \ddot X \hi j |  U \hi {ij}}-  \Sigma \lo {\ddot X \hi i \ddot X \hi j |  U \hi {ij}} \| \looo {hs} = O \lo P (b \lo n).
\end{align*}
\end{theorem}

\medskip

Finally,  we combine  Theorem \ref{theorem:ccco approximation 1} through Theorem \ref{theorem:ccco approximation 2} to come up with  the convergence rate of $
\hat \Sigma \lo {\ddot X \hi i \ddot X \hi j | \hat U \hi {ij}}$. Since there are numerous cross references among the conditions in these theorems, to make a clear presentation we list all the original conditions in the next theorem, even if they already appeared. These conditions are of two categories: those for the step 1 that involves sufficient dimension reduction of $X \hi {(i,j)}$ versus $X \hi {-(i,j)}$, and those for the step 2 that involves the estimation of the conjoined conditional covariance operator. We refer to them  as the first-level and second-level conditions, respectively.

\begin{theorem}\label{theorem:final ccco rate} Suppose the following conditions hold:
\begin{enumerate}
\item[(a)] (First-level kernel) $E [\ka   (S, S)] < \infty$ for $\ka = \ka \lo X \hi {(i,j)}$ and $\ka = \ka \lo X \hi {-(i,j)}$;
\item[(b)] (First-level operator)  $\Sigma \lo {X \hi {-(i,j)} X \hi {(i,j)}}$ is a finite-rank operator with rank $d \lo {ij}$ and
\begin{align*}
\ali \ran ( \Sigma \lo {X \hi {-(i,j)} X \hi {(i,j)} } ) \subseteq \ran ( \Sigma \lo {X \hi {-(i,j)} X \hi {-(i,j)} } \hi {\red{2}}), \\
\ali \ran ( \Sigma \lo {X \hi {(i,j)} X \hi {-(i,j)} } ) \subseteq \ran ( \Sigma \lo {X \hi {(i,j)} X \hi {(i,j)} });
\end{align*}
all the  nonzero eigenvalues of  $ \Sigma \lo {X \hi {(i,j)} X \hi {-(i,j)} } \Sigma \lo {X \hi {-(i,j)} X \hi {-(i,j)} }\inv \Sigma \lo {X \hi {-(i,j)} X \hi {(i,j)} }$ are distinct;
\item[(c)] (First-level tuning parameters) $n \hi {-1/2} \prec \eta \lo n \prec 1$, $n \hi {-1/2} \prec \epsilon  \lo n \prec 1$, $\eta \lo n \hi {-3/2} \epsilon \lo n \hi {-1}  n \hi {-1} + \eta \lo n \hi {-1} n \hi {-1/2} +  \eta \lo n \hi {1/2} + \epsilon \lo n \prec 1$;
\item[(d)] (Second-level kernel) $E [\ka   (S, S)] < \infty$ is satisfied for $\ka = \ka \lo U \hi {ij}$, $\ka \lo {XU} \hi {i,ij}$, and $\ka \lo {XU} \hi {j,ij}$; furthermore, they are transparent kernels;
   \item[(e)] (Second-level operators) $
\Sigma \lo {U \hi {ij}U \hi {ij}} \inv \Sigma \lo {U \hi {ij}(X \hi i U \hi {ij})}$ and $ \Sigma \lo {U \hi {ij}U \hi {ij}} \inv \Sigma \lo {U \hi {ij}(X \hi j U \hi {ij})}
$
are bounded linear operators;
\item[(f)] (Second-level tuning parameter)  $\delta \lo n \asymp \eta \lo n \hi {-3/2} \epsilon \lo n \hi {-1}  n \hi {-1} + \eta \lo n \hi {-1} n \hi {-1/2} +  \red{\eta \lo n} + \epsilon \lo n$.
\end{enumerate}
Then
\begin{align}\label{eq:final convergence rate}
\| \hat \Sigma \lo {\ddot X \hi i \ddot X \hi j | \hat U \hi {ij}}- \Sigma \lo {\ddot X \hi i \ddot X \hi j |   U \hi {ij}} \| \looo {hs} = O \lo P (\eta \lo n \hi {-3/2} \epsilon \lo n \hi {-1}  n \hi {-1} + \eta \lo n \hi {-1} n \hi {-1/2} +  \red{\eta \lo n} + \epsilon \lo n).
\end{align}
\end{theorem}

\medskip

Using this result we immediately arrive at the variable selection consistency of the Sufficient Graphical Model.

\begin{corollary} Under the conditions in Theorem \ref{theorem:final ccco rate}, if
\begin{align*}
\ali \eta \lo n \hi {-3/2} \epsilon \lo n \hi {-1}  n \hi {-1} + \eta \lo n \hi {-1} n \hi {-1/2} +  \red{\eta \lo n } + \epsilon \lo n \prec \rho \lo n \prec 1, \ \mbox{ and} \\
\ali \hat {\ca E} = \{ (i,j) \in \Gamma \times \Gamma: \ i > j, \ \|  \hat \Sigma \lo {\ddot X \hi i \ddot X \hi j | \hat U \hi {ij}} \| \looo {hs} < \rho \lo n \}
\end{align*}
then $\lim \lo {n \to \infty} P ( \hat {\ca E} = \ca E ) \to 1$.
\end{corollary}

\subsection{Optimal rates of tuning parameters}

The convergence rate in Theorem \ref{theorem:final ccco rate} depends on $\epsilon \lo n$ and $\eta \lo n$ explicitly, and $\delta \lo n$ implicitly (in the sense that $\delta \lo n \asymp
\eta \lo n \hi {-3/2} \epsilon \lo n \hi {-1}  n \hi {-1} + \eta \lo n \hi {-1} n \hi {-1/2} +  \red{\eta \lo n } + \epsilon \lo n$ is optimal for fixed $\epsilon \lo n$ and $\eta \lo n$). Intuitively, when $\epsilon \lo n$, $\eta \lo n$, and $\delta \lo n$ increase, the biases increase  and variances decrease; when they decrease, the biases decrease  and the variances increase. Thus there should be critical rates for them that balance the bias and variance, which are the optimal rates.

\begin{theorem}\label{theorem:optimal tuning}  Under the conditions in Theorem \ref{theorem:final ccco rate}, if $\epsilon \lo n$, $\eta \lo n$, and $\delta \lo n$ are of the form $n \hi a$, $n \hi b$, and $n \hi c$ for some $a > 0$, $b > 0$, and $c > 0$, then
\begin{enumerate}
\item[(i)] the optimal rates the tuning parameters are
\begin{align*}
n \hi {-\red{3/8}} \preceq \epsilon \lo n \preceq n \hi {-\red{1/4}}, \quad \eta \lo n \asymp n \hi {-\red{1/4}}, \quad \delta \lo n \asymp n \hi {-\red{1/4}};
\end{align*}
\item[(ii)] the optimal convergence rate of the estimated conjoined conditional covariance operator is
\begin{align*}
\| \hat \Sigma \lo {\ddot X \hi i \ddot X \hi j | \hat U \hi {ij}}- \Sigma \lo {\ddot X \hi i \ddot X \hi j |   U \hi {ij}} \| \looo {hs} = O \lo P (n \hi {-\red{1/4}}).
\end{align*}
\end{enumerate}
\end{theorem}

\medskip

Note that there is a range of $\epsilon \lo n$ are optimal, this is because the convergence rate does not have a unique minimizer. This also means the result is not very sensitive to this tuning parameter.

\blue{In the above asymptotic analysis, we have treated $p$ as fixed when $n \to \infty$. We have also developed the consistency and convergence rate in the scenario where the dimension of $p \lo n$ of $X$ goes to infinity with $n$, which is placed in the Supplementary Material (Section \rm S9) due to limited space. }

\section{Simulation}\label{sec;simulation}
In this section we compare the performance of our sufficient graphical model  with previous methods such as \citet{yuan2007model}, \citet{liu2009nonparanormal}, \citet{voorman2013graph}, \cite{fellinghauer2013stable}, \citet{lee2016additive}, \blue{ and a Na\"ive method which is based on the conjoined conditional covariance operator without the dimension reduction step}.

By design,  the sufficient graphical model has advantages over these existing methods under the following circumstances. First,  since the sufficient graphical model does not make any distributional assumption, it should  outperform \citet{yuan2007model} and \citet{liu2009nonparanormal} when the Gaussian or copula Gaussian assumptions are violated; second, due to the sufficient dimension reduction in sufficient graphical model, it   avoids the curse of dimensionality and  should outperform  \citet{voorman2013graph}, \cite{fellinghauer2013stable}, and a Na\"ive method in the high-dimensional setting; third, since sufficient graphical model does not require additive structure, it should outperform \citet{lee2016additive} when there is severe nonadditivity in the model. Our simulation comparisons will reflect these aspects.


For the sufficient graphical model, \citet{lee2016additive}, \blue{and the Na\"ive} method, we use the Gaussian radial basis function as the kernel. The regularization constants $\epsilon\lo X\hi{(i,j)}$, $\epsilon\lo X\hi{-(i,j)}$, and $\epsilon\lo U\hi{(i,j)}$ are chosen by the generalized cross validation criterion described in Section \ref{sec;tuning} with the grid  $\{10\hi{-\ell}: \ell=-1,0,1,2,3,4\}$. The kernel parameters  $\gamma \lo X \hi {(i,j)}$, $\gamma \lo X \hi {-(i,j)}$, $\gamma \lo {XU} \hi {i,ij}$, $\gamma \lo {XU} \hi {j,ij}$, and $\gamma \lo U \hi {ij}$ are chosen according to  (\ref{eq:tuning gamma}). Because the outcomes of tuning parameters are stable, for each model, we compute the generalized cross validation for the first five samples and use their average value for the rest of the simulation.
The performance of each estimate is assessed using the averaged receiver operating characteristic curve as a function of the threshold $\rho$. 
The accuracy of a method across all $\rho$ is measured by the area under the receiver operating characteristic curve.

To isolate the factors that affect accuracy, we first consider two models with relatively small  dimensions and large sample sizes, which are 

\begin{align*}
\text{Model \rom{1}}: \quad  X\hi1 \ali =\epsilon\lo1, \  X\hi2=\epsilon\lo2, \  X\hi3=\text{sin}(2X\hi1)+\epsilon\lo3\\
X\hi4 \ali = (X\hi1)\hi2 + (X\hi2)\hi2+\epsilon\lo4, \  X\hi5=\epsilon\lo5,  \\
\text{Model \rom{2}}: \quad  X\hi1 \ali =\epsilon\lo1, \  X\hi2=X\hi1+ \epsilon\lo2, \  X\hi3=\epsilon\lo3, \ X\hi4 = (X\hi1 + X\hi3)\hi2+\epsilon\lo4, \\
X\hi5 \ali =\text{cos}(2X\hi2X\hi3)+\epsilon\lo5, \  X\hi6=X\hi4+\epsilon\lo6,
\end{align*}

where $\epsilon\lo i$, $i=1, \dots, p$ are from independent and identically distributed standard normal distribution. The edge sets of the two models are 

\begin{align*}
\mbox{Model \rom{1}}: \ali \quad \ca E = \{ (1,3), (1, 4), (2,4), (1,2)\}  \\
\mbox{Model \rom{2}}: \ali \quad \ca E = \{(1,2), (1,4), (3, 4), (1,3), (2,5), (3, 5), (2, 3), (4, 6) \}.
\end{align*}

\begin{figure}[]
\centering
\begin{subfigure}{.49\textwidth}
\centering
\includegraphics[width = 1.4in]{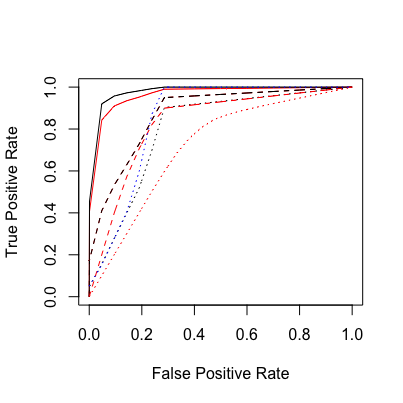} \hspace{-.1in}
\includegraphics[width = 1.4in]{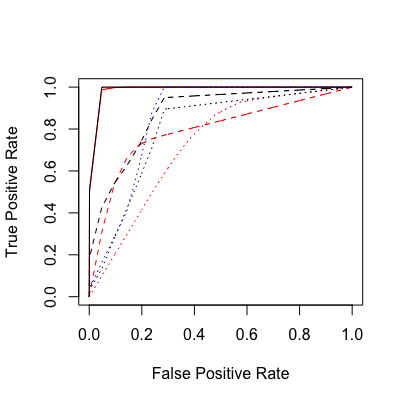}
\caption{Model \RN{1}}
\label{model1}
\end{subfigure}
\begin{subfigure}{.49\textwidth}
\centering
\includegraphics[width = 1.4in]{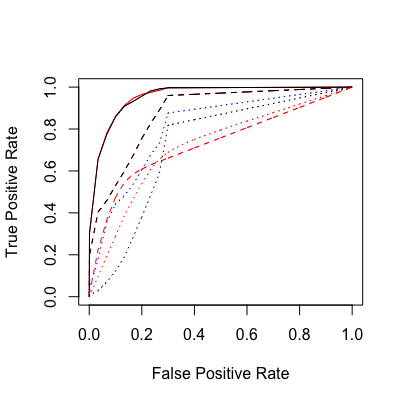} \hspace{-.1in}
\includegraphics[width = 1.4in]{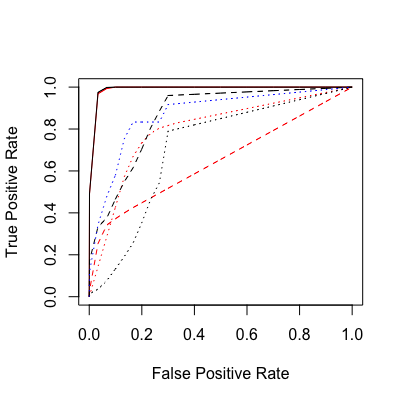}
\caption{Model \RN{2}}
\label{model2}
\end{subfigure}
\centering
\begin{subfigure}{.49\textwidth}
\centering
\includegraphics[width = 1.4in]{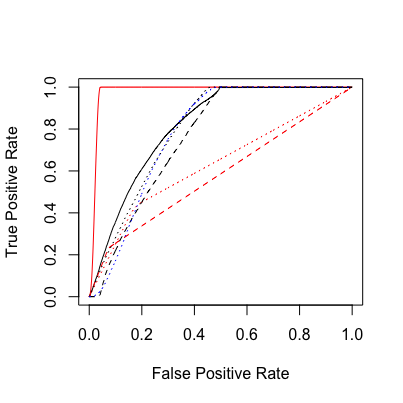} \hspace{-.1in}
\includegraphics[width = 1.4in]{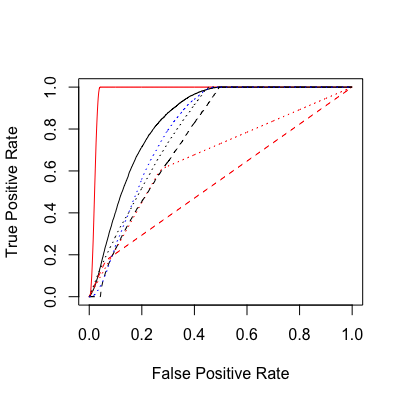}
\caption{Model \RN{3}}
\label{model3}
\end{subfigure}
\begin{subfigure}{.49\textwidth}
\centering
\includegraphics[width = 1.4in]{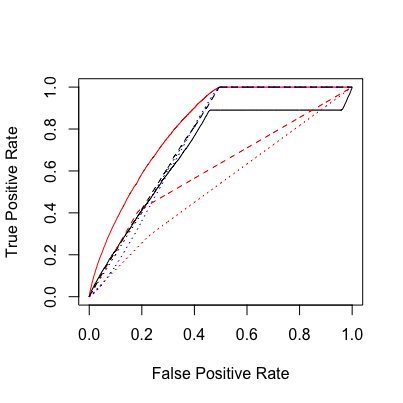} \hspace{-.1in}
\includegraphics[width = 1.4in]{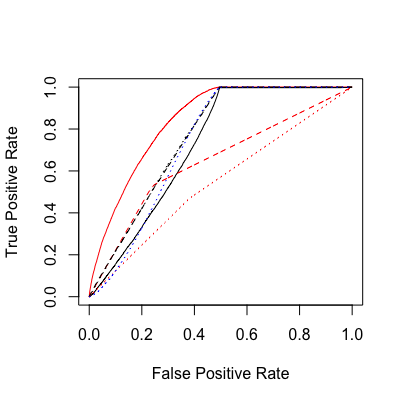}
\caption{Model \RN{4}}
\label{model4}
\end{subfigure}
\caption{Averaged receiver operating characteristic curves of four models. For Model \RN{1} and \RN{2}. Left panel: $n=100$; right panel: $n=1000$.  For \text{Model \rom{3} and \RN{4}}, Left panel: $n=50$; right panel: $n=100$.}
\label{model1rocd2}
\end{figure}

We use $n = 100, 1000$ for each model, and for each $n$, we generate 50 samples to compute the averaged receiver operating characteristic curves. The dimension $d \lo {ij}$ for sufficient graphical model is taken to be 2 for all cases (we have also used $d \lo {ij} = 1$ and the results are very similar to those presented here).
The plots in the first row of Figure \ref{model1rocd2} show the averaged receiver operating characteristic curves for the seven methods, with the following plotting symbol assignment:

\begin{center}
\begin{tabular}{llll}
Sufficient graphical model: & red solid line     \hspace{.05in}  &    \cite{voorman2013graph}: & red dotted line      \\
\cite{lee2016additive}: & black solid line    &   \cite{fellinghauer2013stable}: & black dotted line           \\
\cite{yuan2007model}:  & red dashed line    &    \blue{Na\"ive:} & \blue{blue dotted line}           \\
\cite{liu2009nonparanormal}: & black dashed line     &&
\end{tabular}
\end{center}

From these figures we see that
the two top performers are clearly sufficient graphical model and \cite{lee2016additive}, and their performances are very similar. Note that none of the two models satisfies the Gaussian or copula Gaussian assumption, which explains why sufficient graphical model and \cite{lee2016additive} outperform \cite{yuan2007model} and \cite{liu2009nonparanormal}. Sufficient graphical model and \cite{lee2016additive} also outperform \cite{voorman2013graph}, \cite{fellinghauer2013stable}, \blue{and Na\"ive method}, indicating that  curse of dimensionality already takes effect on the fully nonparametric methods. \blue{The three nonparametric estimators have similar performances.} Also note that Model I has an additive structure, which explains the slight advantage of \cite{lee2016additive} over sufficient graphical model in subfigure (a) of Figure \ref{model1rocd2}; Model II is not additive, and the advantage of \cite{lee2016additive} disappears in subfigure (b) of Figure \ref{model1rocd2}.

We next consider two models with relatively high dimensions and small sample sizes. A convenient systematic way to generate   larger networks is via the hub structure. We choose $p = 200$, and randomly generate ten hubs  $h \lo 1, \ldots, h \lo {10}$ from  the 200 vertices. For each $h \lo k$, we randomly select a set $H \lo k$ of 19 vertices  to form  the neighborhood of $h \lo k$. With the network structures thus specified, our two probabilistic models are
\begin{align*}
\ali  \text{Model \RN{3}}:   \quad  X\hi i = 1+ \vert X\hi{h\lo k} \vert\hi2 + \epsilon\lo i, \quad \text{where} \quad i \in H\lo k \setminus h\lo k,   \\
\ali      \text{Model \RN{4}}:  \quad  X\hi i =  \sin((X\hi{h\lo k})\hi3) \epsilon\lo i, \quad \text{where} \quad i \in H\lo k \setminus h\lo k,
\end{align*}
\blue{and $\epsilon \lo i$'s are the same as in  Models \RN{1} and \RN{2}.} Note that, in Model III, the dependence of $X \lo i$ on $X \lo {h \lo k}$ is through the conditional mean $E ( X \lo i | X \lo {h \lo k})$, whereas in Model IV, the dependence is through the conditional variance $\var ( X \lo i | X \lo {h \lo k })$.
For each model, we choose two sample sizes $n=50$ and $n=100$. The averaged receiver operating characteristic curves (again averaged over 50 samples) are presented in the second row in Figure \ref{model1rocd2}. From the figures we see that, in the high-dimensional setting with \kyongwon{$p > n$}, sufficient graphical model substantially outperforms all the other methods, which clearly indicates the benefit of  dimension reduction in constructing graphical models.

\begin{figure}[]
\centering
\includegraphics[width =2in]{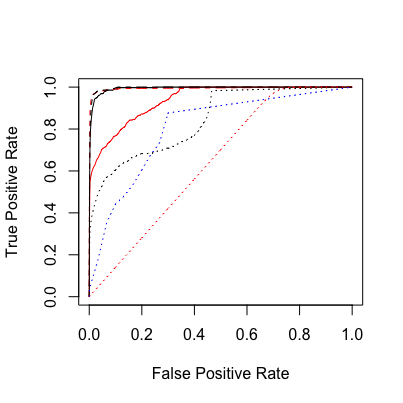}\quad
\includegraphics[width =2in]{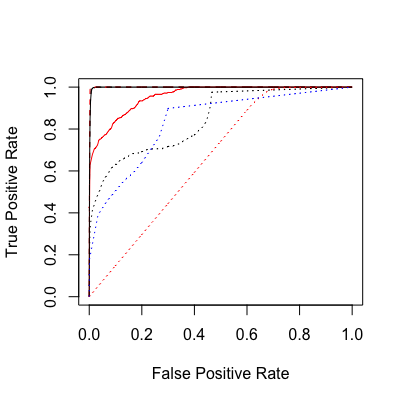}
\caption{Averaged receiver operating characteristic curves for \text{Model \rom{5}}. Left panel: $n=100$; right panel: $n=200$. }
\label{gaussiand2}
\end{figure}

\blue{We now consider a Gaussian graphical  model  to investigate any efficiency loss incurred by sufficient graphical model. Following the similar structure used in \cite{li2014additive}, we choose  $p=20$, $n=100, 200$, and the model
\begin{align*}
 \text{Model \RN{5}}:    X \sim N(0, \Theta\hi{-1}),
\end{align*}
where $\Theta$ is $20 \times 20$ precision matrix with diagonal entries 1, 1, 1, 1.333, 3.010, 3.203, 1.543, 1.270, 1.544, 3, 1, 1, 1.2, 1, 1, 1, 1, 3, 2, 1, and nonzero off-diagonal  entries $\theta\lo{3,5}=1.418$, $\theta\lo{4,10}=-0.744$, $\theta\lo{5,9}=0.519$, $\theta\lo{5,10}=-0.577$, $\theta\lo{13,17}=0.287$, $\theta\lo{17,20}=0.542$, $\theta\lo{14,15}=0.998$. As expected, Figure \ref{gaussiand2} shows that \cite{yuan2007model}, \cite{liu2009nonparanormal}, and \cite{lee2016additive} perform better than sufficient graphical model in this case. However, sufficient graphical model still performs reasonably well and significantly outperforms the fully nonparametric methods.
}

\blue{Finally, we conducted some simulation on the generalized cross validation criterion (\ref{eq:GCV for rho}) for determining the threshold $\rho \lo n$. We generated samples from  Models I through V as described above, produced the receiver operating characteristic curves using sufficient graphical model, and determined the threshold $\rho \lo n$ by (\ref{eq:GCV for rho}). The results are presented in Figure \rm S1 in the Supplementary Material. In each penal, the generalized cross validation-determined threshold $\rho \lo n$ are represented by the black dots on the red receiver operating characteristic curves.  }

\section{Application}\label{sec;application}
We now apply sufficient graphical model to a data set from the DREAM 4 Challenge project and compare it with other methods.
The goal of this Challenge is to recover gene regulation networks from simulated steady-state data.
A description of this data set can be found in \citet{marbach2010revealing}.
Since \citet{lee2016additive} already compared  their method with \cite{yuan2007model}, \cite{liu2009nonparanormal}, \cite{voorman2013graph}, \cite{fellinghauer2013stable}, \blue{and Na\"ive method} for this dataset and demonstrated the superiority of \cite{lee2016additive} among these estimators, here we will focus on the comparison of the sufficient graphical model with \cite{lee2016additive} and the champion method for the DREAM 4 Challenge.

The data set contains data from five networks each  of dimension of 100 and sample size 201. We use the Gaussian radial basis function kernel for sufficient graphical model and \cite{lee2016additive} and the tuning methods described in Section \ref{sec;tuning}. For sufficient graphical model, the dimensions $d \lo {ij}$ are taken   to be 1. We have also experimented with  $d \lo {ij} = 2$ but the results (not presented here) show no significant difference. Because networks are available, we can compare the receiver operating characteristic curves and their areas under the curve's, which are shown in Table \ref{apptable}.

\begin{table}[H]
\caption{Comparison of sufficient graphical model, \cite{lee2016additive}, Na\"ive and the champion methods in DREAM 4 Challenge}
\label{apptable}
\begin{tabular}{c c c c c c}

    & Network 1 & Network 2 & Network 3 & Network 4 & Network 5 \\ 
Sufficient graphical model & 0.85      & 0.81      & 0.83      & 0.83      & 0.79      \\ 
\cite{lee2016additive} & 0.86      & 0.81      & 0.83      & 0.83      & 0.77      \\ \
Champion & 0.91      & 0.81      & 0.83      & 0.83      & 0.75      \\ 
\blue{Na\"ive} & \blue{0.78}      & \blue{0.76}      & \blue{0.78}      & \blue{0.76}      & \blue{0.71}      \\  
\end{tabular}
\end{table}

As we can see from Table \ref{apptable}, sufficient graphical model has the same areas under the receiver operating characteristic curve values as \cite{lee2016additive} for  Networks 2, 3, and 4, performs better than \cite{lee2016additive} for Network 5, but trails slightly behind \cite{lee2016additive} for Network 1; sufficient graphical model has the same areas under the curve as the champion method, performs better  for Network 5 and worse for Network 1. Overall, sufficient graphical model and \cite{lee2016additive} perform  similarly in this dataset, and they are on a par with the champion method.  We should point out that  sufficient graphical model and \cite{lee2016additive} are purely empirical; they employ  no knowledge about the underlying physical mechanism generating the gene expression data. However, according to  \citet{pinna2010from}, the champion method did use a differential equation that reflects the underlying physical mechanism.
\blue{The results for threshold determination are presented in Figure \rm S2 in the Supplementary Material.}

\section{Discussion}\label{sec;discussion}

This paper is a first attempt to take advantage of the recently developed nonlinear sufficient dimension reduction method to nonparametrically estimate the statistical graphical model while avoiding the curse of dimensionality. Nonlinear sufficient dimension reduction is used as a module and applied repeatedly to evaluate  conditional independence, which leads to a substantial gain in accuracy in the high-dimensional setting.
Compared with the Gaussian and copula Gaussian methods, our method is not affected by the violation of the Gaussian and copula Gaussian assumptions. Compared with the additive method \citep{lee2016additive}, our method does not require an additive structure and retains the   conditional independence as the criterion to determine the edges, which is a commonly accepted criterion. Compared with fully nonparametric methods,  sufficient graphical model avoids the curse of dimensionality and significantly enhances the performance.

The present framework  opens up several directions for further research. First, the current model assumes that the central class $\frak S \lo {X \hi {(i,j)} | X \hi {-(i,j)}}$ is complete, so that generalized sliced inverse regression is the exhaustive nonlinear sufficient dimension reduction estimate. When this condition is violated, generalized sliced inverse regression is no longer exhaustive and we can employ other nonlinear sufficient dimension reduction methods such as the generalized sliced averaged variance estimation \citep{lee2013general,li2018sufficient} to recover the part of the central class that generalized sliced inverse regression misses.  Second,  though we have assumed that there is a proper sufficient sub-$\sigma$-field $\ca G \hi {-(i,j)}$ for each $(i,j)$, the proposed estimation procedure is still justifiable when no such sub-$\sigma$-field exists. In this case,    $U \hi {ij}$ is still the most important set of functions that characterize the statistical dependence of $X \hi {(i,j)}$ on $X \hi {-(i,j)}$ -- even though it is not sufficient. Without sufficiency, our method may be more appropriately called the Principal Graphical Model  than the sufficient graphical model. Third, the current method can be extended to functional graphical model, which are common in medical applications such as EEG and fMRI. Several functional graphical models have been proposed recently, by
\citet{zhu2016Bayesian}, \citet{qiao2019functional}, \citet{li2018nonparametric}, and \citet{solea2020copula}. The idea of a sufficient graph can be applied to this setting to improve efficiency.

This paper also contains some theoretical advances that are novel to nonlinear sufficient dimension reduction. For example, it introduces a general framework to characterize how the error of nonlinear sufficient dimension reduction propagates to the downstream analysis in terms of convergence rates. \blue{Furthermore, the results for convergence rates of various linear operators allowing the dimension of the predictor to go to infinity are the first of its kind in nonlinear sufficient dimension reduction. These advances will benefit the future development of sufficient dimension reduction in general, beyond the current context of estimating graphical models.}

\acks{Bing Li's research on this work was supported in part by   the NSF Grant DMS-1713078. Kyongwon Kim's work was supported by the National Research Foundation of Korea(NRF) grant funded by the Korea government(MSIT) (No.2021R1F1A1046976, RS-2023-00219212), basic Science Research Program through the National Research Foundation of Korea(NRF) funded by the Ministry of Education (2021R1A6A1A10039823).}

\section*{Supplementary Material}
Supplementary material includes proofs of all theorems, lemmas, corollaries, and propositions in the paper, asymptotic development for the high-dimensional setting, some additional simulation plots for threshold determination.

\vskip 0.2in
\bibliography{bibliography}

\end{document}